\documentclass[runningheads]{llncs}
\usepackage[T1]{fontenc}
\usepackage{graphicx}
\usepackage{cite}
\usepackage{amsfonts}
\usepackage{amsmath}
\usepackage{xcolor}
\usepackage{algorithm}
\usepackage{algpseudocode}
\usepackage{subcaption}

\algrenewcommand\algorithmicrequire{\textbf{Input:}}
\algrenewcommand\algorithmicensure{\textbf{Output:}}
\algnewcommand{\LineComment}[1]{\State \(\triangleright\) #1}
\usepackage{adjustbox}
\usepackage{bm}
\usepackage[title]{appendix}

\begin{document}
\pretolerance=10000
\title{ECATS: Explainable-by-design concept-based anomaly detection for time series}
\titlerunning{ECATS} 
% If the paper title is too long for the running head, you can set
% an abbreviated paper title here

\author{Irene Ferfoglia\inst{1}\orcidID{0000-0003-1585-6576} \and
Gaia Saveri\inst{1,2}\orcidID{0009-0003-2948-7705} \and \\
Laura Nenzi\inst{1}\orcidID{0000-0003-2263-9342} \and
Luca Bortolussi\inst{1}\orcidID{0000-0001-8874-4001}}
\authorrunning{I. Ferfoglia et al.}
% First names are abbreviated in the running head.

\institute{Università degli Studi di Trieste, Italy \and Università degli Studi di Pisa, Italy}% \and
% Springer Heidelberg, Tiergartenstr. 17, 69121 Heidelberg, Germany
%\email{lncs@springer.com}\\
%\url{http://www.springer.com/gp/computer-science/lncs} \and
%ABC Institute, Rupert-Karls-University Heidelberg, Heidelberg, Germany\\
%\email{\{abc,lncs\}@uni-heidelberg.de}}
%
\maketitle 
\begin{abstract} 
Deep learning methods for time series have already reached excellent performances in both prediction and classification tasks, including anomaly detection. However, the complexity inherent in Cyber Physical Systems (CPS) creates a challenge when it comes to explainability methods.
To overcome this inherent lack of interpretability, we propose ECATS, a concept-based neuro-symbolic architecture where concepts are represented as Signal Temporal Logic (STL) formulae. Leveraging kernel-based methods for STL, concept embeddings are learnt in an unsupervised manner through a cross-attention mechanism. 
The network makes class predictions through these concept embeddings, allowing for a meaningful explanation to be naturally extracted for each input.
Our preliminary experiments with simple CPS-based datasets show that our model is able to achieve great classification performance while ensuring local interpretability.

\end{abstract}

\section{Introduction}
Cyber-physical systems (CPS) are commonly used for the monitoring and controlling of industrial processes in important assets such as power grids, water treatment facilities, and autonomous vehicles. The monitoring of the sensor readings in such systems is critical for early detection of unexpected system behaviour as it allows for timely attenuating actions - such as fault checking, predictive maintenance, and system shutdown - to be taken, such that potential problem caused by system failures can be avoided. The increasing complexity of modern CPS has made traditional anomaly detection mechanisms insufficient. Consequently, the use of machine learning and deep learning algorithms has become a clear direction to build data-driven anomaly detection frameworks.

The signals of a CPS are mainly time series data that are continuously recorded over time. While deep learning models have been able to correctly predict the presence of an anomaly in time series \cite{darban2022deep}, most methods do so in a black-box manner. Lack of transparency in the decision process is a characteristic of many deep learning models. This creates a deficiency of human trust despite their state-of-the-art performance across most tasks \cite{LEICHTMANN2023107539}, raising ethical \cite{ethics1, ethics2} and legal \cite{legal, gdpr} concerns. Specifically, anomaly detection is critical for maintaining CPS' integrity and functionality and can greatly benefit from an interpretable model to guide the work. 

Indeed, explainability is now an important research topic in the field of AI \cite{electronics12051092}. Historically, a large part of the work in explainability is done on tabular data \cite{9551946} or in the field of computer vision \cite{buhrmester2021analysis}, where deep neural networks (DNN) typically achieve state-of-the-art performance. The research field of eXplainable AI (XAI) for trajectory classification has seen a surge of attention since  2019 \cite{xaitimeseries, rojat2021explainable}. Common explainability methods for time series are based on Convolutional Neural Networks (CNN), articulating in backpropagation-based methods \cite{zhou2015learning, wang2016time, Strodthoff_2019} and perturbation-based methods \cite{kashiparekh2019convtimenet}. % The former provide explanations by doing a single forward and backward pass in the network and include methods such as class activation mapping (CAM) \cite{zhou2015learning, wang2016time} and "Gradient*input" \cite{Strodthoff_2019}, while the latter directly compute the contribution of the input features by altering them, running a forward pass and measuring the difference with the original input \cite{ancona2018better}, such as with the occlusion sensitivity method used in \cite{kashiparekh2019convtimenet}. \\
The other common approach for XAI methods in time series is coming from Natural Language Processing (NLP) and is rooted in the attention mechanism %, which assign values corresponding to the importance of different parts of the inputs and can be calculated in many different ways 
\cite{Karim_2018, vinayavekhin2018focusing, ge2018interpretable}.

On the other hand, concept-based models are designed to increase explainability in deep learning models by exploiting more human-understandable concepts to describe their decision process. This can be done using methods such as logistic regression or decision trees ~\cite{pmlr-v119-koh20a, kazhdan2020cme} or, as the latest state-of-the-art, concept embeddings. However, currently, most of these methods focus on the classification of images \cite{img1, img2}, tabular data \cite{tab1}, or graphs \cite{dcr-icml23}, leaving a gap to fill in terms of trajectory data. 

In this paper, we propose ECATS, the first interpretable neuro-symbolic concept-based model for trajectories. It uses a cross-attention-based network to perform binary classification on some potentially anomalous trajectories by leveraging STL formulae as concepts and exploiting their properties.

\section{Preliminaries}
\subsubsection{Concept-based models}
Recent work on explainable AI introduces this novel type of explanation approach \cite{ghorbani2019automatic, NEURIPS2020_ecb287ff}, providing human-understandable concepts as explanations for the predictions rather than the relative importance of input features, better resonating with human reasoning.
Explainable-by-design concept-based models depart from standard neural networks by explicitly incorporating a set of concepts within the architecture, which can be provided as part of the dataset or computed by the model itself. % They can either employ a dataset with annotated concepts or without concepts annotations. 

\vspace{-0.5cm}
\subsubsection{Signal Temporal Logic}\label{stl} 
Signal Temporal Logic (STL) is a formalism used in formal verification and synthesis of CPS. STL formulae describe relationships between signals over dense time intervals, allowing us to express properties such as "the temperature should never exceed a certain threshold for more than a specified duration". Besides classical propositional logic operators, STL syntax is endowed with \emph{temporal} operators eventually, globally and until, decorated with time intervals $[a,b]$. They can be intuitively interpreted as: a property is \textit{eventually} satisfied if it is satisfied at some point inside the temporal interval, while a property is \textit{globally} satisfied if it is true continuously in $[a, b]$; finally the \textit{until} operator captures the relationship between two conditions $\varphi, \psi$ in which the first condition $\varphi$ holds until, at some point in $[a, b]$, the second condition $\psi$ becomes true. Notably, STL is endowed with both \emph{qualitative} (or Boolean) semantics, giving the classical notion of satisfaction of a property over a trajectory, and 
a \emph{quantitative} semantics (or \emph{robustness}) $\rho(\varphi, \xi)$, which is a measure of how robust is the satisfaction of $\varphi$ w.r.t. perturbations of the signal $\xi$. A formal definition of STL syntax and semantics is given in Appendix~\ref{app:stl}.

\vspace{-0.4cm}
\subsubsection{A kernel for STL formulae} Finding continuous representations of STL formulae is performed with an ad hoc kernel in \cite{stl-kernel}, by leveraging the quantitative semantics of STL. Indeed, robustness allows formulae to be considered as functionals mapping trajectories into real numbers, i.e.\ $\rho(\varphi,\cdot): \mathcal{T}\rightarrow \mathbb{R}$ such that $\xi\mapsto \rho(\varphi, \xi)$. Considering these as feature maps, and fixing a probability measure $\mu_0$ on the space of trajectories $\mathcal{T}$, a kernel function capturing similarity among STL formulae can be defined as:
\begin{equation}
k(\varphi, \psi) = \langle \rho(\varphi, \cdot), \rho(\psi, \cdot) \rangle = \int_{\xi\in \mathcal{T}} \rho(\varphi, \xi) \rho(\psi, \xi) d\mu_0(\xi)
\label{eq:stl-kernel}
\end{equation}
opening the doors to the use of the scalar product in the Hilbert space $L^2$ as a kernel for $\mathcal{P}$, the space of parameters. Intuitively, this results in a kernel having a high positive value for formulae that behave similarly on high-probability trajectories (w.r.t. $\mu_0$) and, vice versa, low negative value for formulae that on those trajectories disagree. Finite-dimensional embeddings can be recovered by kernel PCA~\cite{kpca}. More details about the kernel can be found in Appendix~\ref{app:kernel}.

\section{ECATS architecture}
Here we introduce ECATS (Explainable-by-design Concept-based Anomaly detection model for Time Series), the first interpretable neuro-symbolic concept-based model for trajectories. The goal of our proposed methodology is to discriminate between normal and anomalous signals while simultaneously providing a contextual explanation for the detected outcome. The form of the explanation in ECATS is a disjunction of interpretable Signal Temporal Logic(STL) formulae, which function as concepts in our model, and whose semantics influence the labeling of the input trajectories. As mentioned in Section~\ref{stl}, STL is a concise yet expressive language suited for describing the characteristics of time series data, making it a valuable candidate to serve as a concept language in this field. In fact, small STL formulae describe simple patterns in a signal, and are easy to understand for a human (possibly once converted into a natural language description).

\begin{figure}[t]
  \centering
  \includegraphics[width=0.65\linewidth]{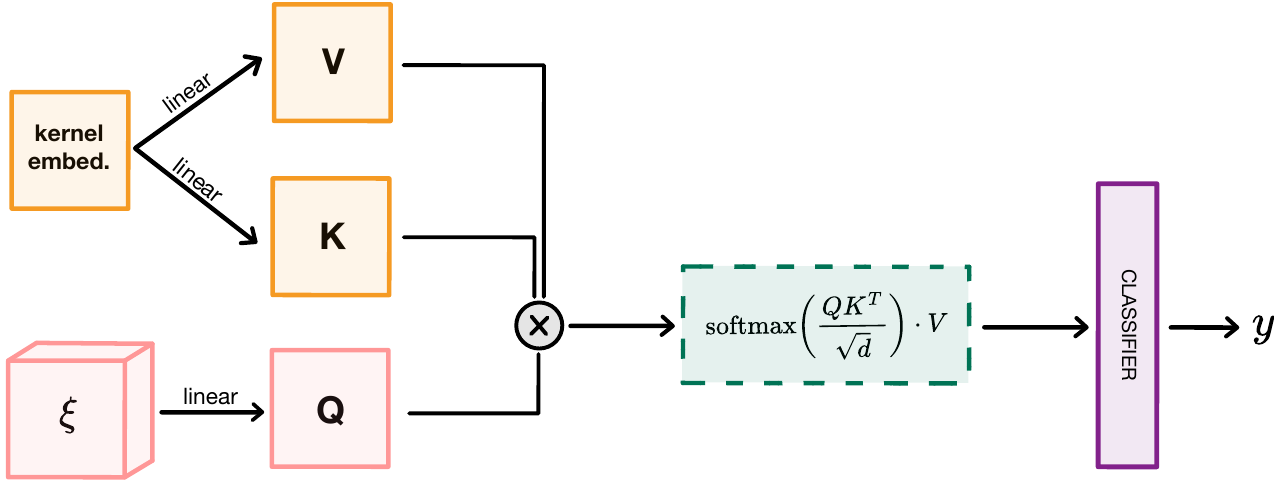}
  \caption{Architecture of ECATS. $V, K, Q$ are respectively the values, keys and query matrices typical of an attention mechanism.} 
  \label{fig:model}
\end{figure}

We now describe at a high level how the model works. We provide Figure~\ref{fig:model} as a reference to graphically follow the discussion. Our model takes as input a signal $\xi$ and, through a cross-attention mechanism (Subsection~\ref{crossatt}) between a kernel embedding of the STL-based concepts (computed as per Equation~\ref{eq:stl-kernel}) and the input trajectory, creating an efficient and detailed representation of their relationship. The representation  % Then, the incorporation of a truth degree element (Subsection~\ref{tdeg}), in blue in Figure~\ref{fig:model}, creates an efficient representation of the relationship between concepts and trajectories. Lastly, this 
is then passed to an Multi-Layer Perceptron (MLP) classifier network, which outputs its prediction $y$ indicating the class of the input trajectory. Our model is then able to provide not only excellent classification accuracy (Section~\ref{exp}), but also transparent, easy-to-read explanations made of simple STL formulae (Subsection~\ref{expl}).

\subsection{Cross-attention} 
\label{crossatt}
Since its introduction, self-attention \cite{vaswani2023attention} has been widely used in many different forms to enhance feature representations. By simulating how human attention works and assigning varying levels of importance to different parts of the input, a model implementing attention is able to weight the importance of each element in the input concerning all other elements. \\
While self-attention captures relationships within a single input sequence, cross-attention \cite{chen2021crossvit} captures the relationships between elements of two different input sequences, allowing the model to generate coherent and contextually relevant outputs. Applied to our case, cross-attention allows us to take a trajectory and compute its attention values with respect to a kernel embedding of a fixed dataset of STL formulae. We selected the set of formulae $\mathcal{C}$ to be used as concepts with a procedure aimed at maximising the coverage of the semantic space of the formulae and avoiding semantic redundancies, namely sampling from the latent space defined by the STL kernel and inverting the embeddings to identify the corresponding formulae (in fact, simple formulae with a very close embedding). For more details, see Appendix~\ref{app:concepts}. The use of cross-attention produces an attention matrix which will be used also when retrieving an explanation for the output of our model, other than for prediction.

% \subsection{Concept truth degree}
% \label{tdeg}
% Before being fed through the classifier, the product of the cross-attention mechanism is summed with a nonlinear transformation of the \textit{concept truth degrees}. These are the robustness degrees of each STL formula in the dataset, computed at time $0$ for each trajectory in the input. For more information on robustness and how it is computed, we refer to Appendix~\ref{app:stl}.

\subsection{Concept-based explanation} \label{expl}
\paragraph{Local explanations} can be extracted for the output of each classified trajectory. Given a set of concepts $\mathcal{C}$ identified as described above, we aim to explain the model prediction for a given input trajectory $\xi$. The explanation can be expressed as a subset of $\mathcal{C}$, i.e. $\mathcal{E} \subseteq \mathcal{C}$, where each formula $\varphi \in \mathcal{E}$ is associated with its robustness $\rho(\varphi, x)$ w.r.t. the input trajectory and its attention weight computed through a forward pass. Robustness helps determine the role that $\varphi$ played in the classification (i.e. if they are positive or negative witnesses). For more information on the robustness, see Appendix~\ref{app:stl}. \\ % The sign of these values represent the truthfulness of a formula on a trajectory and therefore the \textit{role} (positive/negative) of the concept in the classification process. \\
The formulae are then ranked by their attention weight and can be filtered by kernel similarity (see Appendix~\ref{app:concepts}) to obtain a more meaningful subset of formulae. The final explanation is the conjunction of the contents of the subset.

\paragraph{Global explanations} by class, can be easily extracted from local ones. In general, each local explanation may have different weights and concepts. However, we can still globally interpret the predictions of our model without the need for an external post-hoc explainer. To achieve this, we collect the batch of unfiltered formulae produced as explanation for each trajectory of a class with the highest attention value. We then filter this batch by kernel similarity to finally obtain a handful of STL formulae able to characterise the class they refer to.

\paragraph{A post-processing step} is applied to improve human readability and comprehension. While the model may be able to distinguish between classes with a given formula, the separation may not be intuitive to the human eye. So, a given formula's threshold is shifted in a way that it becomes true for a class, and false for the other, i.e. opposite signs of robustness. If the formula is given an explanation for a trajectory or a class for which it is false, the formula gets negated in order to invert the roles and become true for the element in focus. They can be condensed into a single formula by disjunction. See Appendix~\ref{app:simplify} for details.

\section{Experiments}
\label{exp}
\subsubsection{Setup} 
\begin{figure}[b]
    \centering
    \includegraphics[width=0.6\linewidth]{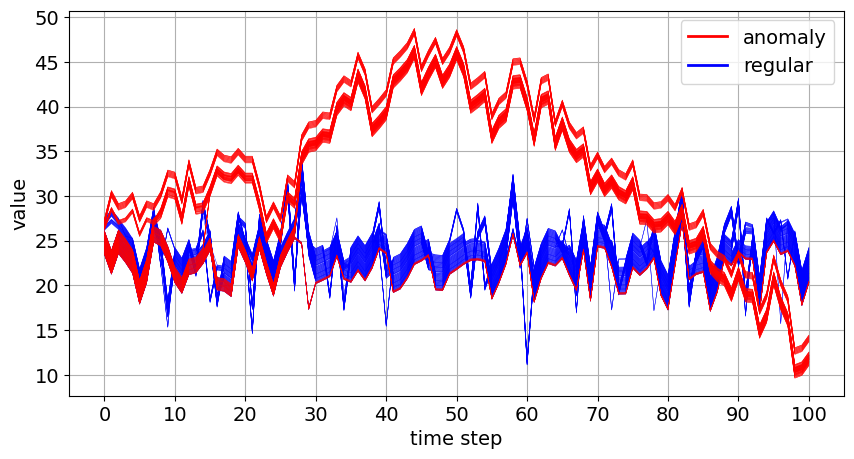}
    \caption{Train cruise control trajectories. Blue trajectories are regular, red trajectories are anomalous.}
    \label{fig:dataset}
\end{figure}

We evaluate ECATS on a commonly-used CPS-based dataset, a train cruise control dataset, often used as learning benchmark \cite{bustle, train2, train3}. The dataset, balanced between classes, collects $200$ trajectories of $1$ dimension and has a clear-cut condition to classify between regular and anomalous trajectories, except for seven outliers. See Figure~\ref{fig:dataset} for a visualisation of the trajectories of the dataset, with the anomalies highlighted in red.

\subsubsection{Evaluation}\label{exp-eval}
To assess the classification performance of our model, we run the tests using $5$ different initialisation seeds. \\

For the train cruise control dataset, the classification accuracy averages at $96.5\%$, with a standard deviation of $2.55$, not reaching perfection most likely due to the outliers. Following, some example outputs from ECATS, starting from a local explanation.

\begin{figure}[t]
\centering
\begin{minipage}[t]{.34\textwidth}
  \centering
  \includegraphics[width=1\linewidth]{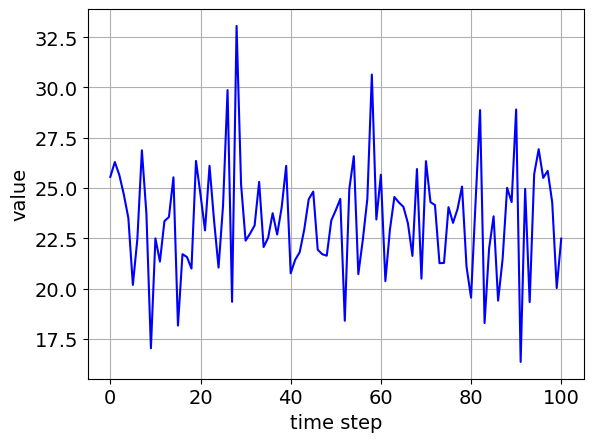}
  \captionof{figure}{The regular trajectory in the train cruise control dataset considered as example.}
  \label{fig:regular1v}
\end{minipage}%
\qquad
\begin{minipage}[t]{.6\textwidth}
  \centering
\includegraphics[width=\linewidth]{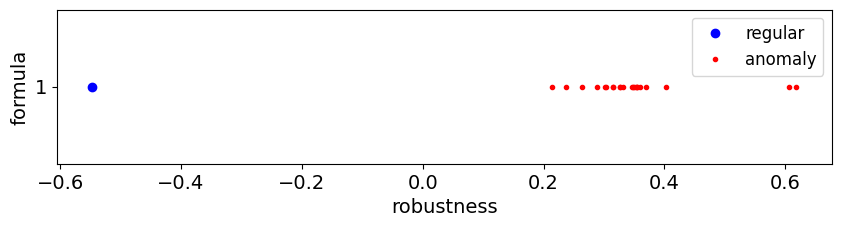}
\caption{Robustness values for the explanation of the regular trajectory reported as example.}
\label{fig:normalrob}
\end{minipage}
\end{figure}

Consider the post-processed explanation given for a signal $\xi$ correctly classified as regular, which can be visualised in Figure~\ref{fig:regular1v}: 
\begin{equation*}
    \neg \, (\xi \le 35.9 \ \mathcal{U}_{[11,36]} \ \xi \ge 31.5)
\end{equation*}
The first part of the formula, $\xi \le 35.9$, is true for almost all trajectories, no matter the class. However, the second part, $\xi \ge 31.5$, is not true for regular classes, making the formula inside the parentheses false for the considered signal. This is backed by the robustness of this STL formula on the trajectory, namely $-0.55$. With the negation, its robustness becomes positive and the formula becomes true for the regular class and helps indeed discriminate between regular and anomalous trajectories while characterising the studied signal. \\
We further analyse this result by reporting the plot of the robustness values for the considered signal versus all the signals of the opposite class (Figure~\ref{fig:normalrob}). 

As per Figure~\ref{fig:normalrob}, the provided formula does an excellent job at separating the signal belonging to the regular class from the ones that are anomalous. Coincidentally the cluster of values regarding the anomalous class has opposite sign with respect to the robustness on the considered regular signal, therefore the formula has no need to be shifted to increase human readability. \\

\begin{figure}[t]
\centering
\begin{minipage}[t]{.34\textwidth}
  \centering
    \includegraphics[width=\linewidth]{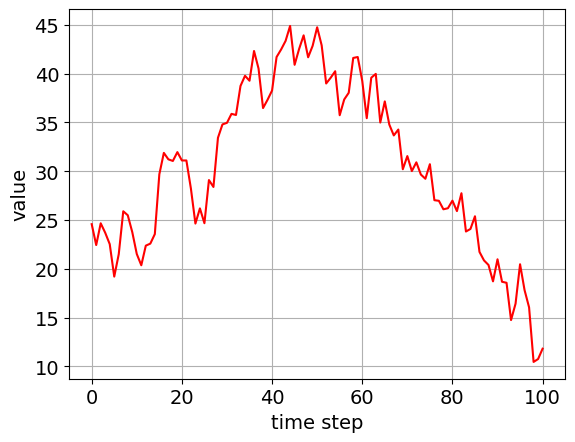}
    \caption{The anomalous trajectory in the train cruise control dataset considered as example.}
    \label{fig:anomaly}
    \vfill
\end{minipage}%
\qquad
\begin{minipage}[t]{.6\textwidth}
  \centering
\includegraphics[width=\linewidth]{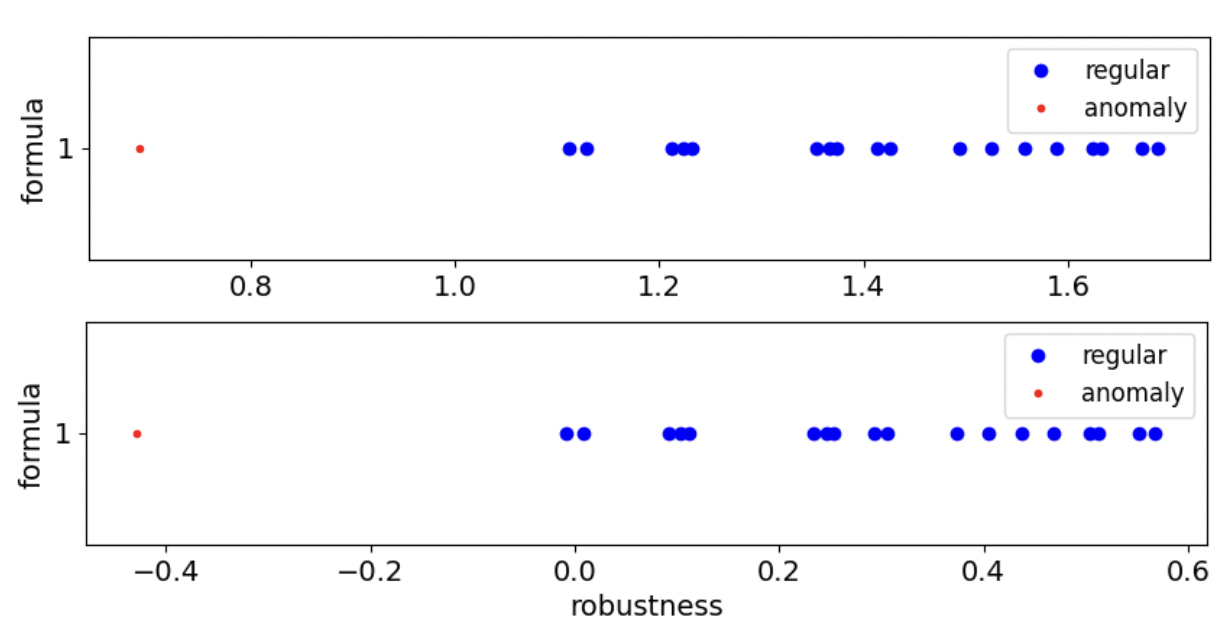}
\caption{Robustness values for the explanation of the anomalous trajectory reported as example, before (top) and after (bottom) postprocessing.}
\label{fig:anomalyrob}
\end{minipage}
\end{figure}

However, this is not always the case. We now report a local explanation for an anomalous trajectory, represented in Figure~\ref{fig:anomaly}. The original provided explanation is the following. 
\begin{equation*}
    \mathcal{G}_{[0,24]}  \ \xi \le 37.3
\end{equation*}
 Indeed, also from Figure~\ref{fig:dataset} one can assess that the formula is true for all trajectories in the dataset. This is not an intuitive explanation, but we can see from the corresponding robustness plot (Figure~\ref{fig:anomalyrob}, top) that it does in fact separate the two classes. So, we can shift the values in the formula, obtaining the following post-processed explanation and robustness values of inverse sign (Figure~\ref{fig:anomalyrob}, bottom):
 \begin{equation*}
    \neg \, \mathcal{G}_{[0,24]}  \ \xi \le 28.7
\end{equation*}
See Appendix~\ref{app:simplify} for more details on the simplification of formulae. \\

\begin{figure}
    \centering
    \includegraphics[width=0.8\linewidth]{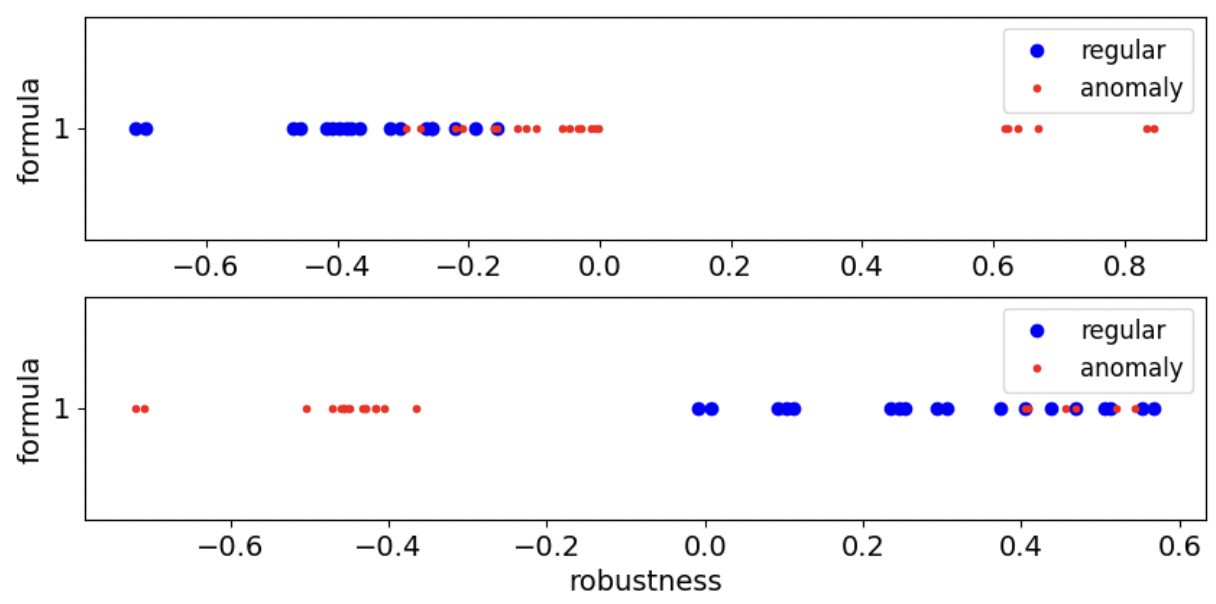}
     \caption{Robustness of the ECATS's global explanation for regular (top) and anomalous (bottom) trajectories’ class.}
    \label{fig:glob}
\end{figure}

Lastly, here is an example of post-processed global explanations given by ECATS:
\begin{align*}
    \text{Regular:  }&\neg \, (\xi \ge 20.0 \ \mathcal{U}_{[23,47]} \ \xi \le 40.5) \\
    \text{Anomaly:  }&\neg \, \mathcal{G}_{[0,24]}  \ \xi \le 28.7
\end{align*}
In Figure~\ref{fig:glob}, the robustness values computed for all trajectories for each formula provided as explanation, where each point in the plot represents the value for one trajectory. Except for some points which are most likely the aforementioned outliers, the class separation is linear. 

For training details, further discussion of results on the train cruise control dataset, and some preliminary results on a 2-variable dataset, we refer to Appendix~\ref{app:results}. 

\section{Conclusions}
We introduced ECATS, a pioneering interpretable neuro-symbolic concept-based model designed specifically for the classification of trajectory data in CPS. By leveraging STL formulae as concepts and using a cross-attention-based network, we proved that our model is able to provide not only accurate anomaly detection but also transparent and human-understandable explanations, giving insight into the decision-making processes. 
This not only facilitates timely and appropriate responses to potential system failures but also promotes greater user confidence in the model's predictions. 

A significant strength of our approach is its versatility in various practical scenarios. Thanks to ECATS' ability to provide not only global but also local explanations, it can be effectively applied even in cases where: i) a labeled dataset is not available, ii) only semi-supervised data is present, iii) there is minimal prior knowledge about the data, or iv) the amount of data is limited. This flexibility makes ECATS a robust solution for a wide range of real-world applications where data constraints and uncertainties are common.

\paragraph{Future work} will aim to extend this model capabilities to multivariate trajectories and investigate the potential integration of additional explainability techniques to further enhance its utility and effectiveness. Lastly, further automatic post-processing of the explanations is needed in order to provide even more clear and understandable STL formulae.

\begin{credits}
\subsubsection{\ackname}
This study was carried out within the PNRR research activities of the consortium iNEST (Interconnected North-Est Innovation Ecosystem) funded by the European Union Next-GenerationEU (Piano Nazionale di Ripresa e Resilienza (PNRR) – Missione 4 Componente 2, Investimento 1.5 – D.D. 1058 23/06/2022, ECS\_00000043) and the MUR PRIN project 20228FT78M DREAM (modular software design to reduce uncertainty in ethics-based cyber-physical systems). This manuscript reflects only the Authors’ views and opinions, neither the European Union nor the European Commission can be considered responsible for them.

% \subsubsection{\discintname}

\end{credits}

\bibliographystyle{splncs04}
\bibliography{paper}

\begin{subappendices}
\renewcommand{\thesection}{\Alph{section}}%
% or try \arabic{section}

\section{Signal Temporal Logic}\label{app:stl}
Signal Temporal Logic (STL) \cite{stl} is a formal specification language to express temporal properties over real-value trajectories with a dense time interval. With STL, one can state in a rich but compact way properties of signals that vary over time. For example, in STL we can express properties like "For the next 3 days the highest temperature will be below 75 degree and the lowest temperature will be above 60 degree" or "There must exist a time point in the next 10 time units, at which the velocity of the car will be less than 2". We now present its syntax and semantics. 
% An \textit{atomic proposition} $\mu$ is an inequality of the form $y(x) > 0 \ (y : \mathbb{R}^n \to \mathbb{R})$, where $x : \mathcal{X} \to \mathbb{R}^n$ is a trajectory. The syntax of an STL formula is defined as follows: $$ \varphi:=\top \, |\, \mu \, |\, \neg\varphi \,|\, \varphi_1 \land \varphi_2 \, |\, \phi_1 \,\mathcal{U}_I \, \varphi_2$$

For the rest of the paper, let be $\xi : \mathcal{T} \to \mathbb{R}^n$ a trajectory, where $\mathcal{T} = \mathbb{R}\ge0$ is the time domain, $\xi_i (t)$ is the value at time $t$ of the projection on the $i$-th coordinate, and $\xi = (\xi_1 , ..., \xi_n )$, as an abuse on the notation, is used also to indicate the set of variables of the trace considered in the formulae.
\begin{definition}
The syntax of STL is given by $$ \varphi:=\top \, |\, \mu \, |\, \neg\varphi \,|\, \varphi_1 \land \varphi_2 \, |\, \phi_1 \,\mathcal{U}_I \, \varphi_2$$
where $\top$ is the Boolean true constant, negation $\neg$ and conjunction $\land$ are the standard Boolean connectives, and $\mathcal{U}_I $ is the Until temporal modality, where $I$ is a real positive interval. We can derive the disjunction operator $\lor$, the future eventually $\mathcal{F}_\mathcal{I} $ and always $\mathcal{G}_\mathcal{I}$ operators from the until temporal modality $(\varphi_1 \lor \varphi_2 = \neg(\neg\varphi_1 \land \neg\varphi_2), \ \mathcal{F}_\mathcal{I} \varphi = \top\mathcal{U}_\mathcal{I} \varphi, \ \mathcal{G}_\mathcal{I} \varphi = \neg\mathcal{F}_\mathcal{I} \neg \varphi)$. \\

\end{definition}

Referring to the example mentioned above, we can translate them from plain English to STL as follows: "For the next 3 days the highest temperature will be below 75 degree and the lowest temperature will be above 60 degree" is $G_{[0, 3]}(T<75 \wedge T>60)$, while "There must exist a time point in the next 10 time units, at which the velocity of the car will be less than 2" is $F_{[0, 10]} (v < 2)$.

STL can be interpreted over a trajectory $\xi$ using a qualitative or a quantitative semantics \cite{stlrob}. The quantitative satisfaction function $\rho$ returns a value $\rho(\varphi, x, t) \in \bar{\mathbb{R}}$ \footnote{$\bar{\mathbb{R}} =\mathbb{R} \cup \{ - \infty , + \infty\} $} quantifying the \textit{robustness degree} of the property $\varphi$ by the trajectory $\xi$ at time $t$. Its sign provides the link with the standard Boolean semantics in that a signal $\xi$ satisfies an STL formula $\varphi$ at a time $t$ iff the robustness degree $\rho(\varphi,\xi,t) \ge 0$. Its absolute value, instead, can be interpreted as a measure of the robustness of the satisfaction with respect to noise in signal $\xi$, measured in terms of the maximal perturbation in the secondary signal $\theta(\xi(t))$, preserving truth value. 
% add definition of robustness
Robustness is recursively defined as:
\begin{align*}
 & \rho(\pi,\xi,t) &=& f_\pi(\xi(t)) \qquad \text{for } \pi(\bm{x})=\big(f_\pi(\bm{x})\geq 0\big)\\
 & \rho(\lnot\varphi,\xi,t) &=& -\rho(\varphi,\xi,t)\\
 & \rho(\varphi_1\land\varphi_2,\xi,t) &=& \min\big(\rho(\varphi_1,\xi,t), \rho(\varphi_2,\xi,t)\big)\\
 &\rho(\varphi_1\mathbf{U}_{[a, b]}\varphi_2,\xi,t) \hspace*{-0.5em}&=& \max_{{t'\in[t+a,t+b]}}\big(\min\big(\rho(\varphi_2,\xi,t'),
 \min_{{t''\in[t,t']}}\rho(\varphi_1,\xi,t'')\big)\big) 
\end{align*}
For completeness, we report also the definition of robustness of derived temporal operators: eventually $\rho(\mathbf{F}_{[a, b]}\varphi, \xi, t) = \max_{t'\in[t+a,t+b]} \rho(\varphi,\xi,t)$ and globally $\rho(\mathbf{G}_{[a, b]}\varphi, \xi, t) = \min_{t'\in[t+a,t+b]} \rho(\varphi,\xi,t)$. Robustness is compatible with satisfaction via the following \emph{soundness} property: if $\rho(\varphi, \xi, t) > 0$ then $s(\varphi, \xi, t) = 1$ and if $\rho(\varphi, \xi, t) < 0$ then $s(\varphi, \xi, t) = 0$. When $\rho(\varphi, \xi, t) = 0$ arbitrary small perturbations of the signal might lead to changes in satisfaction value. We omit $t$ from the previous notation when the properties are evaluated at time $t=0$. 

\section{STL Kernel}\label{app:kernel}
In what follows, we provide details on the measure $\mu_0$ used in the space of trajectories, to get a more intuitive understanding of the kernel for STL formulae defined in Section~\ref{stl} (and, in particular, Equation~\ref{eq:stl-kernel}). 

\begin{algorithm}[H]
\caption{Sampling a trajectory over the interval $[a, b]$ according to $\mu_0$}
\label{alg:mu0}
\begin{algorithmic}
% \LineComment $\Delta$ discretization step
\Require $\Delta$, $a$, $b$, $m'$, $m''$, $\sigma'$, $\sigma''$, $q$
\Ensure $\xi$ 
\LineComment sample the starting point 
\State $\xi_0\sim\mathcal{N}(m', \sigma')$
\State $\xi(t_0) \gets \xi_0$
\LineComment sample the total variation
\State $K\sim(\mathcal{N}(m'', \sigma''))^2$
\State $y_1,...,y_{N-1}\sim\mathbb{U}([0, K])$
\State $y_0\gets0$, $y_n\gets K$
\State orderAndRename($y_0, \ldots, y_n$) 
\LineComment now $y_1 \leq y_2 \leq...\leq y_{N-1}$
\State $s_0\sim\text{Discr}(-1, 1)$
\While{$i\leq N$}
\State $s\gets$Binomial($q$) \Comment $P(s=-1) = q$
\State $s_{i+1} = s_i\cdot s$ 
\State $\xi(t_{i+1}) = \xi(t_i) + s_{i+1}(y_{i+1}-y_i)$ 
\EndWhile
\end{algorithmic}
\end{algorithm}

Intuitively, $\mu_0$ makes \emph{simple} trajectories more probable, considering total variation and number of changes in monotonicity as indicators of complexity of signals. The measure $\mu_0$, operating on piecewise linear functions over the interval $\mathcal{I} = [a, b]$ (which is a dense subset of the set of continuous functions over $\mathcal{I}$), can be algorithmically defined as in Algorithm~\ref{alg:mu0} (default parameters are set as $a=0, b=100, \Delta=1, m'=m''=0'', \sigma'=\sigma''=1, q=0.1$).
Note that although the feature space $\mathbb{R}^{\mathcal{T}}$ into which $\rho$ (and thus Equation (\ref{eq:stl-kernel})) maps formulae is infinite-dimensional, in practice the kernel trick allows to circumvent this issue by mapping each formula to a vector of dimension equal to the number of formulae which are in the training set used to evaluate the kernel (Gram) matrix.

\section{Selecting the concepts}\label{app:concepts}
We construct the STL formulae used as concepts by fixing the maximum number $M$ of nodes and the maximum number of variables $N$ (i.e. the dimensionality of the input signals) allowed, then we enumerate all templates (i.e., STL formulae where constants are replaced by parameters) satisfying those constraints as detailed in Algorithm~\ref{alg:stl-templates}.

\begin{algorithm}[t]
\begin{adjustbox}{minipage=\linewidth,scale=0.95}
\caption{Algorithms for generating STL formulae templates}
\label{alg:stl-templates}
\begin{algorithmic}
\Require $M$, $N$
\Ensure all\_phis \Comment{\text{STL templates of formulae with max $N$ vars and $M$ nodes}}
\State $\text{all\_phis}\gets []$
\State all\_phis.append(generateAtomicPropositions()) \Comment{$x_i\leq 0$ or $x_i\geq 0$, $\forall i\leq N$}
\For{$2\leq m\leq M$}
\State $\text{prev\_phis}\gets$getPhisGivenNodes($m-1$) \Comment{retrieve templates with $m-1$ nodes}
\State $\text{unary\_ops}\gets$expandByUnaryOperators(prev\_phis) \Comment{$F$, $G$, $\neg$}
\State all\_phis.append(unary\_ops)
\State l\_list, r\_list $\gets$getPairsGivenSum($m$) \Comment{all pairs $(l, r): l+r=m, l\leq r$}
\For{$(l, r)\in$ [l\_list, r\_list]}
\State l\_phis$\gets$getPhisGivenNodes($l$)
\State r\_phis$\gets$getPhisGivenNodes($r$)
\State binary\_ops$\gets$expandByBinaryOpeators(l\_phis, r\_phis) \Comment{$\wedge$, $\vee$, $U$}
\State all\_phis.append(binary\_ops)
\EndFor
\EndFor
\end{algorithmic}
\end{adjustbox}
\end{algorithm}
Once all the templates have been generated, we instantiate each of them on a set of parameter configurations $\mathcal{P} = \{p_0,\ldots p_{|\mathcal{P}|}\}$ to be filtered by Signature-based Optimisation as defined in \cite{enumerate-stl}. Denoting as $\varphi(p)$ the application of the parameter set $p$ to the template formula $\varphi$, and given a set of trajectories $\hat{\mathcal{T}} = \{\xi_0,\ldots, \xi_{|\hat{\mathcal{T}}|}\}$ (with $p\in \mathcal{P}$, and $\hat{\mathcal{T}}\subset \mathcal{T}$), it consists in building a matrix $S\in \mathbb{R}^{|\mathcal{P}|\times |\hat{\mathcal{T}}|}$, whose rows are called the \textit{signature} of the formula, s.t. $S(i, j) = \rho(\varphi(p_i), \xi_j)$, i.e. the cell $(i, j)$ of such matrix holds the robustness of the $i$-th instantiation of $\varphi$ on the $j$-th trajectory. In such matrix, the distance between rows can be considered as a proxy for semantic similarity among different parameterisations of the same template, hence a selection on formulae can be done by keeping only those whose signature has a distance higher than a certain threshold $\tau\in\mathbb{R}_{\geq 0}$ from formulae already selected. This helps in removing redundancy in the set of concepts. In our case, we use signature-based optimisation with cosine distance and threshold $\tau = 0.9$ and, for interpretability, we keep $M$ fairly small, namely $M=3$. Hence, with such procedure, we select as concepts only a subset of all the possible instantiations of each template, following semantic diversity as criterium. Once signature-based optimization is done on all templates, we have the full set of concepts, from which we obtain concept vectors by computing their kernel embeddings.

\subsection{Latin Hypercube Sampling}
Latin Hypercube Sampling (LHS) was used in order to sample a meaningful subset of concepts of size $m$ from their kernel embedding. LHS is a type of stratified Monte Carlo method developed in the 1970s employed to efficiently sample from multidimensional spaces, including matrices. Unlike traditional random sampling methods, LHS ensures a more uniform coverage of the space. By partitioning the range of each component into equally probable intervals and selecting one sample from each interval per component, LHS mitigates the risk of oversampling certain regions while neglecting others, thereby minimising biases inherent in the sampling process. In out case, we selected a pool of $m=5000$ concepts.

\section{Post-processing of explanations}\label{app:simplify} As mentioned in Section~\ref{exp-eval}, although being able to accurately distinguish between classes, STL formulae provided by the model as explanations are not optimised towards their human-understandabilty. In particular, one would expect the robustness of such formulae to be positive (indicating boolean satisfaction) on trajectories belonging to the class they describe, and negative for signals belonging to the opposite class. This can be obtained without altering the discriminative power of the formula by shifting of a fixed quantity $\varepsilon$ all the variable thresholds appearing in the explanation. A trivial way to do so is by trying an arbitrary number of values for $\varepsilon$ (e.g. choosing from a grid of candidates), and keeping the smallest one for which the sign of the robustness is the same for all trajectory belonging to a class, and the opposite for all trajectories belonging to the other. Finally, to achieve the property of having positive robustness on signals of the class pertaining the explanation, one can simply negate the formula. Putting together these post-processing steps transform the original explanation (without changing its separation properties), hence it might be the case that a syntactic simplification is possible, enhancing human readability of the explanation, keeping its semantic untouched (i.e. just applying logical equivalences). Notably, all these post-processing steps can be performed automatically. 

\section{Training setup and results} \label{app:results}
The model was trained for $50$ epochs with a Binary Cross Entropy (BCE) loss and $1e^{-5}$ learning rate, with $5$ different initialisation seeds. We here report and further discuss the results obtained with one of the seeds.

Let's see some other local explanations provided by ECATS, starting from one provided for a regular trajectory.

\begin{figure}[t]
\centering
\begin{minipage}[t]{.34\textwidth}
  \centering
  \includegraphics[width=1\linewidth]{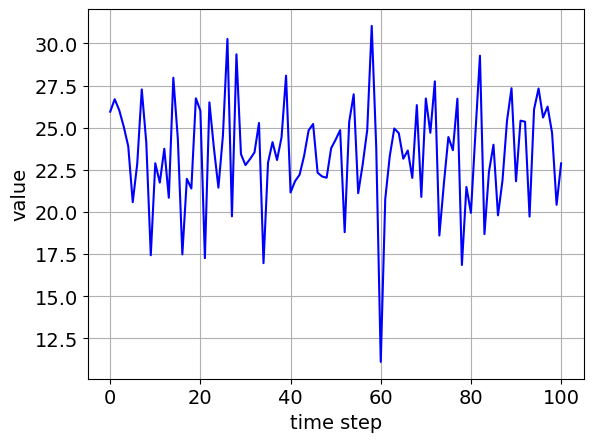}
  \captionof{figure}{The regular trajectory in the train cruise control dataset considered as example in appendix.}
  \label{fig:regular1v2}
\end{minipage}%
\qquad
\begin{minipage}[t]{.6\textwidth}
  \centering
\includegraphics[width=\linewidth]{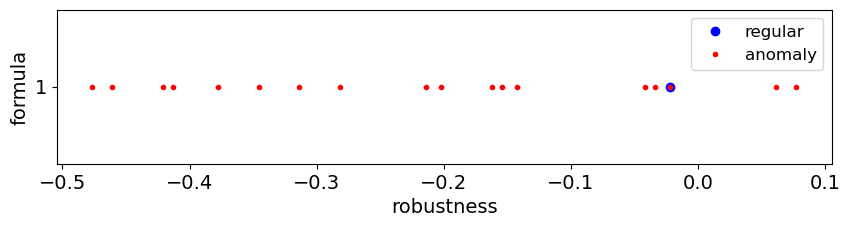}
\caption{Robustness values for the explanation of the regular trajectory considered in appendix.}
\label{fig:normalrob2}
\end{minipage}
\end{figure}

As earlier, consider the post-processed explanation given for a signal $\xi$ correctly classified as regular, which can be visualised in Figure~\ref{fig:regular1v2}: 
\begin{equation*}
     \mathcal{G}_{[0, 24]} \mathcal{F}_{[0,12]} \, \xi \le 37.26
\end{equation*}
Keeping Figure~\ref{fig:dataset} at hand, this formula doesn't seem to describe any particular class since in the considered time interval most trajectories are $\le 37.26$. However, by plotting the robustness (Figure~\ref{fig:normalrob2}) we can see that the formula does discriminate between classes (except for the usual outliers). To aid interpretability, we can shift the threshold, obtaining
\begin{equation*}
     \mathcal{G}_{[0, 24]} \mathcal{F}_{[0,12]} \, \xi \le 27.44
\end{equation*}
Now the formula does indeed discriminate between this regular and the other anomalous trajectories while characterising the studied signal.

\begin{figure}[t]
\centering
\begin{minipage}[t]{.34\textwidth}
  \centering
  \includegraphics[width=1\linewidth]{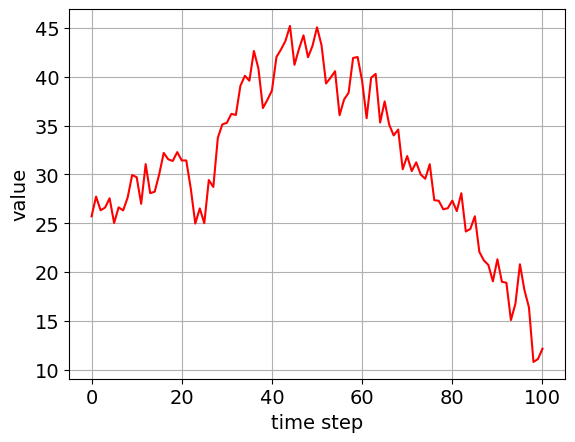}
  \captionof{figure}{The anomalous trajectory in the train cruise control dataset considered in appendix.}
  \label{fig:anomaly2}
\end{minipage}%
\qquad
\begin{minipage}[t]{.6\textwidth}
  \centering
\includegraphics[width=\linewidth]{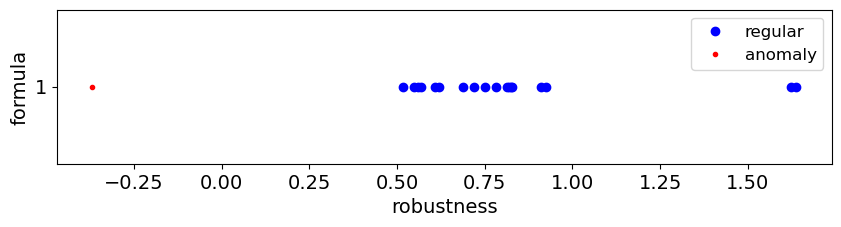}
\caption{Robustness values for the explanation of the anomalous trajectory reported as example in appendix.}
\label{fig:anomalyrob2}
\end{minipage}
\end{figure}

To further prove the capabilities of ECATS, another example of explanation of an anomalous trajectory, reported in Figure~\ref{fig:anomaly2}. The provided explanation for this classification is the following:
\begin{equation*}
     \mathcal{G}_{[0, 36]} 
     \, \xi \ge 37.24
\end{equation*}
As can be seen from the robustness values Figure~\ref{fig:anomalyrob2}, this does discriminate between the considered anomaly and the general regular trajectory. In fact, while at the beginning no trajectories satisfy the condition, towards the last points of the interval the considered anomaly does. This proves the condition to be in fact a good discriminant.

\subsection{Preliminary results with two variables}
\begin{figure}
    \centering
    \begin{subfigure}[b]{0.495\textwidth}
        \centering
        \includegraphics[width=\textwidth]{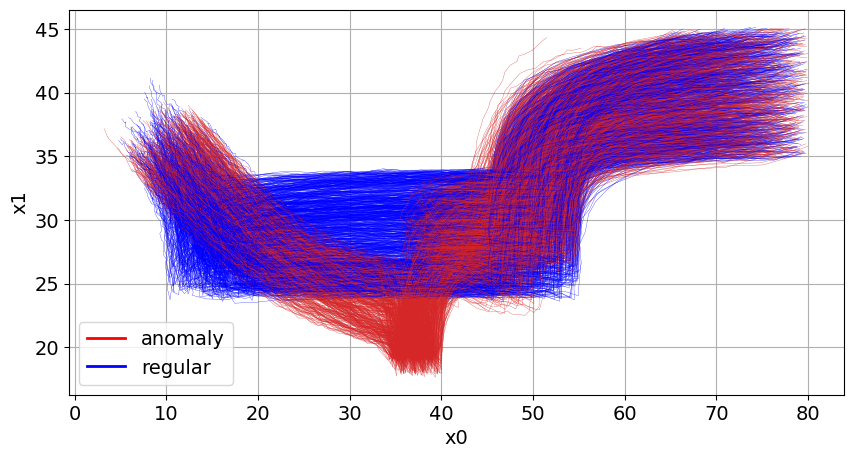}
        \caption{View as plane coordinates.}
        \label{fig:d2notime}
    \end{subfigure}
    \hfill
    \begin{subfigure}[b]{0.495\textwidth}
        \centering
        \includegraphics[width=\textwidth]{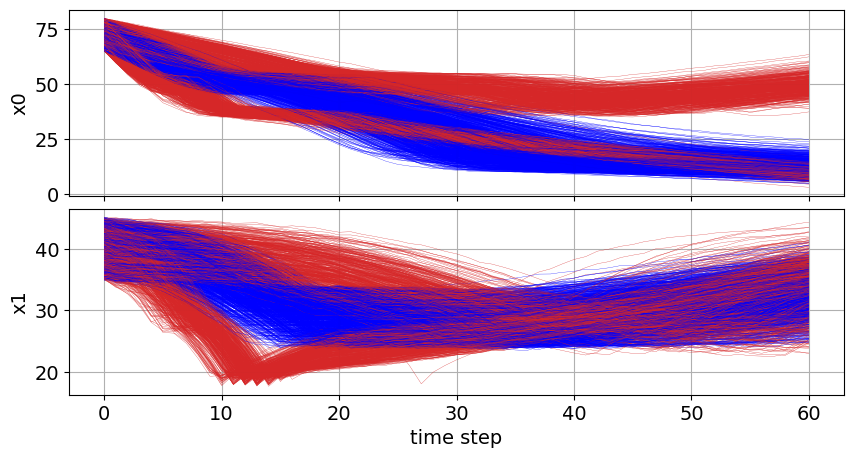}
        \caption{View with reference to time.}
        \label{fig:d2time}
    \end{subfigure}
    \caption{Maritime surveillance trajectories. Blue trajectories are regular, red trajectories are anomalous.}
    \label{fig:dataset2}
\end{figure}

As part of this ongoing research, ECATS has been tentatively applied to a second dataset. This too commonly used \cite{bustle, nenzi2018robust, bombara}, it consists of a maritime surveillance scenario, presenting a more challenging task due to the absence of distinct criteria for distinguishing between regular and anomalous trajectories. The dataset collects $2000$ trajectories, balanced between classes, of $2$ dimensions. Trajectories consist of the longitudinal and latitudinal coordinates ($x_0, x_1$) of vessels transiting over a harbour, with anomalous trajectories corresponding to vessels involved in illegal activities, and regular trajectories corresponding to vessels following the expected route. See Figure~\ref{fig:dataset2} for a visualisation of the trajectories of the dataset, with the anomalies highlighted in red. For better understanding of the explanations, we provide both a view of the variables as coordinates of points on a 2D plane (\ref{fig:d2notime}) and of the value of the single variables related to time (\ref{fig:d2time}).

Note that the following results are preliminary and the model is still to be tweaked to adapt to more complex problems. \\

\begin{figure}
\centering
\begin{minipage}[t]{.34\textwidth}
  \centering
  \includegraphics[width=1\linewidth]{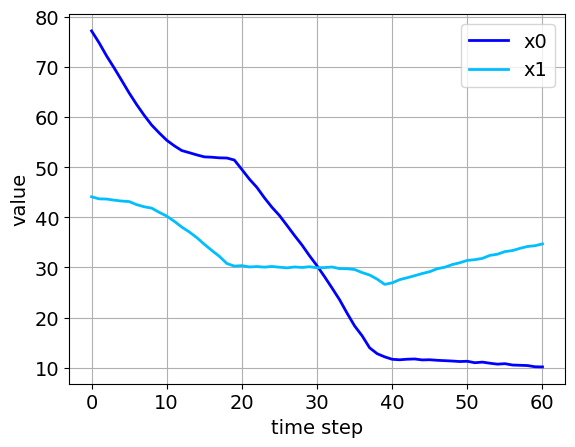}
  \captionof{figure}{The regular trajectory in the maritime dataset considered as example.}
  \label{fig:normal2var}
\end{minipage}%
\qquad
\begin{minipage}[t]{.6\textwidth}
  \centering
\includegraphics[width=\linewidth]{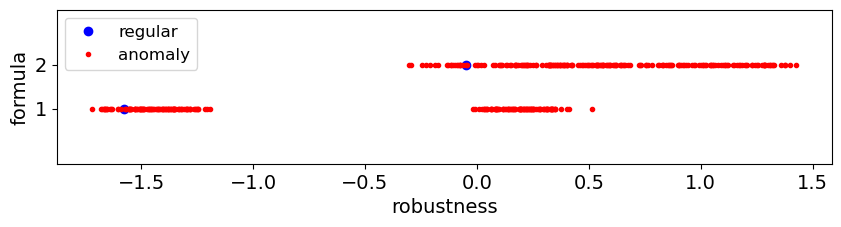}
\caption{Robustness values for the explanation of the regular trajectory.}
\label{fig:normal2rob}
\end{minipage}
\end{figure}

For the maritime dataset, ECATS reaches $100\%$ classification accuracy in all runs. Following, some example outputs for local explanations, starting from a regular trajectory. Consider the post-processed explanation given for a signal $\xi = \{x_0, x_1\} $ correctly classified, which can be visualised in Figure~\ref{fig:normal2var}. The provided explanation is the following:
\begin{equation*}
\mathcal{F}_{21,51]} x_1 \ge 28.38 \ \land \ \mathcal{G}_{[35, \infty]} x_0 \le 18.8 
\end{equation*}

This conjunction of formulas does correctly identify the trajectory. The left one is false for half of the anomalous trajectories, while the right one is false for the other half.

\end{subappendices}
\end{document}